\documentclass[letterpaper, 10 pt, conference]{ieeeconf}  

\IEEEoverridecommandlockouts                              

\usepackage{latexsym}
\usepackage{amsmath}
\usepackage{amssymb}
\usepackage{epsfig}
\usepackage{bbm}
\usepackage{caption}
\usepackage{subcaption}

\usepackage{cite}

\usepackage[ruled]{algorithm2e}

\usepackage{verbatim}
\usepackage{hyperref}
\hypersetup{%
  breaklinks,
  bookmarksnumbered=true,
  bookmarksopen=true,
  colorlinks=true,
  linkcolor=black,
  urlcolor=black,
  citecolor=black
}
\usepackage{xcolor}
\usepackage{subfiles}

\begin{document}
\bstctlcite{IEEEexample:BSTcontrol}
%


\title{
    UNRealNet: Learning Uncertainty-Aware Navigation Features from High-Fidelity Scans of Real Environments
}





\author{
    Samuel Triest$^{1*}$\footnote{work done as an intern at Field AI}, 
    David D. Fan$^{2}$,
    Sebastian Scherer$^{1}$,
    and Ali-Akbar Agha-Mohammadi$^{2}$
    \thanks{* Work done as an intern at Field AI.}%
    \thanks{$^{1}$ Robotics Institute, Carnegie Mellon University, Pittsburgh, PA, USA. \{striest, basti\}@andrew.cmu.edu}%
    \thanks{$^{2}$ Field AI, Mission Viejo, CA, USA.}%
}

\maketitle

%
\IEEEpeerreviewmaketitle

\begin{abstract}


Traversability estimation in rugged, unstructured environments remains a challenging problem in field robotics. Often, the need for precise, accurate traversability estimation is in direct opposition to the limited sensing and compute capability present on affordable, small-scale mobile robots. To address this issue, we present a novel method to learn [u]ncertainty-aware [n]avigation features from high-fidelity scans of [real]-world environments (UNRealNet). This network can be deployed on-robot to predict these high-fidelity features using input from lower-quality sensors. UNRealNet predicts dense, metric-space features directly from single-frame lidar scans, thus reducing the effects of occlusion and odometry error. Our approach is label-free, and is able to produce traversability estimates that are robot-agnostic. Additionally, we can leverage UNRealNet's predictive uncertainty to both produce risk-aware traversability estimates, and refine our feature predictions over time. We find that our method outperforms traditional local mapping and inpainting baselines by up to $\mathbf{40\%}$, and demonstrate its efficacy on multiple legged platforms.


\end{abstract}


\section{Introduction}

More and more robots are being deployed in rough-terrain applications such as construction and disaster response \cite{bellicoso2018advances, afsari2021fundamentals, gehring2021anymal, agha2021nebula, Scherer:2022}. While these robots (which are often legged) have the promise of performing work that is potentially time-consuming and dangerous, these environments often present mobility challenges due to clutter, dangerous terrain, occlusions, and change in the environment over time. As such, there is need for fast, precise and uncertainty-aware traversability analysis. 

At a high level, traversability analysis is the process of constructing cost functions that quantify how desirable it is to reside in a particular region. This process is performed using on-board sensing (such as lidar or camera) and on-board compute. Classical approaches to traversability estimation typically rely on accurate perception to reconstruct the local environment, and apply hand-crafted rules to determine the cost function sent to planning and control modules. Unfortunately, this means that traversability estimation modules are subject to limitations and errors in perception such as odometry drift and occlusion, and often propagate these errors to planning and control, with potentially dangerous consequences for the robot and its operators.

\begin{figure}
    \centering
    \includegraphics[width=0.9\linewidth]{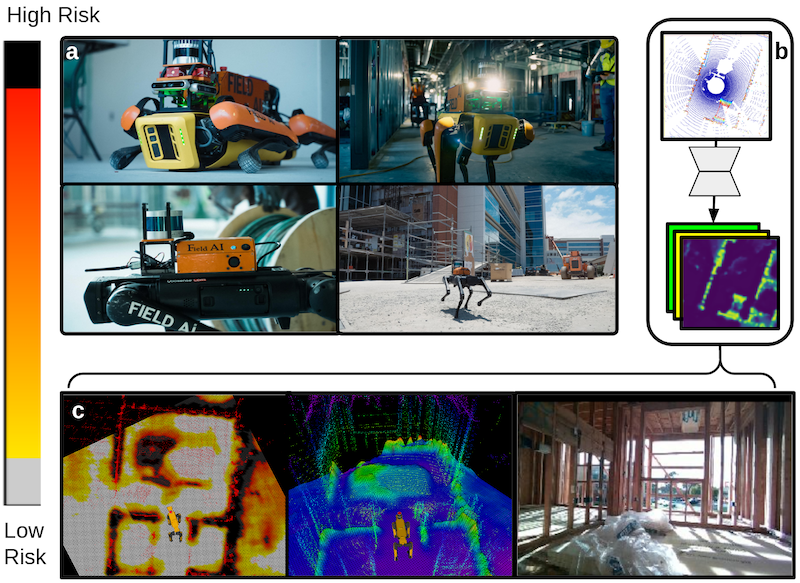}
    \caption{We propose UNRealNet (b), a network that produces high-quality, robot-agnostic navigation features and traversability (c) directly from pointclouds from an on-board sensor (a). Colormap for risk maps included for clarity.}
    \vspace{-0.5cm}
    \label{fig:title_fig}
\end{figure}

Learning presents a promising solution to this problem, as deep neural networks are able to generate complex predictions that capture abstract environmental priors (e.g. most rooms have flat floors, boxes have a particular shape, etc.). However, generating precise ground-truth labels in high quantities to train such a neural network is challenging. Prior work \cite{meng2023terrainnet, gan2022multitask} often leverages aggregation of future sensing (i.e. predict the map at time $t+k$ from the sensing at time $t$), but such approaches are limited by the quality of the SLAM algorithm, the accuracy of the sensor, and the mobility limitations of the robot. The end result is that training labels may be noisy and incomplete.  While simulation may provide a means to address these issues \cite{wang2020tartanair}, simulated environments are often overly simplistic and models trained purely in simulation may not successfully bridge the sim-to-real gap. 


Additionally, uncertainty is a critical component of traversability estimation. In practice, there will be environmental ambiguities (such as occlusion and out-of-distribution objects) that cannot be resolved by on-board sensing. It is thus critical that traversability estimation methods be able to both recognize and address said uncertainty. Learning based-methods have shown promise in uncertainty estimation \cite{kendall2019geometry, georgakis2022uncertainty} for their ability to learn environmental priors, producing uncertainty estimates that go beyond common heuristics.

In this work, we leverage a high-precision laser scanner to generate very accurate and dense reconstructions of several unstructured environments. This allows us to generate high-quality traversability features without the need to address common robotic perception limitations such as occlusion or drift in state estimation. In order to deploy on-robot, we use a label-free learning pipeline where we generate large amounts of simulated lidar data, and the corresponding traversability features. We then train a neural network to predict these features directly from raw pointclouds and demonstrate its efficacy in producing dense, metric traversability estimates directly from sparse lidar data. 

\section{Related Work}

Given its importance in mobile robot autonomy, traversability estimation has been the focus of much research \cite{team2005stanford, jackel2006darpa, papadakis2013terrain, arl_stack}. Of particular importance to this work is the approach described by Fankhauser et al. \cite{fankhauser2018robust} and Chilian et al. \cite{chilian2009stereo}, which is a commonly-used baseline for legged traversability. This method relies on computing several terrain features from a digital elevation map (DEM), which are combined via Equation \ref{eq:fank_trav} into a cost function for downstream planning. For each feature $f$, the user is required to specify two parameters $\alpha_f, f_{crit}$, which specify the importance of that feature, and a critical value over which the robot should not traverse, respectively.

\begin{equation}
    \label{eq:fank_trav}
    c^{i, j} = \sum_f (\alpha_f \frac{f^{i, j}}{f_{crit}})
\end{equation}


While the above methods are well-suited to estimating traversability on fully-observed, accurately-measured terrain, they are brittle with respect to sensing errors and limitations, which are omnipresent in practice. Errors in odometry can lead to terrain surfaces and obstacles being mis-estimated. Dynamic objects such as people and vehicles can cause false positives. Occlusions and negative obstacles can result in no sensing data in areas dangerous to the robot. While methods exist to combat these issues \cite{larson2011lidar, fankhauser2016universal, miki2022elevation}, they are often heuristic, and time-consuming to compute.

Recent research in learning for traversability focuses on learning environment priors to improve traversability estimation in occluded or mis-estimated regions of the environment. Ramakrishnan et al. \cite{ramakrishnan2020occupancy} train a neural network to predict occupancy maps from a ego-centric RGB-D images. They demonstrate this network's ability to leverage environment priors from indoor environments \cite{chang2017matterport3d, xia2018gibson} to improve navigation and exploration. Additional work by Georgakis et al. \cite{georgakis2021learning, georgakis2022uncertainty} extends this idea to semantic mappping and uncertainty-aware exploration. Work done by Fan et al. \cite{fan2021step, fan2021learning} estimate traversability using geometric features of a terrain surface. In contrast to prior work, they define a mapping from terrain features to \textit{distributions} of cost, allowing their navigation behaviors to reason about uncertainty. Stolzle et al. \cite{stolzle2022reconstructing} train a UNet to inpaint elevation maps, given partial elevation maps. They demonstrate that their network is capable of predicting occluded elevation values on real-world data. Gan et al. \cite{gan2022multitask} train a neural network to predict semantics and traversability by building a 3D reconstruction of the environment using the robot's trajectory and running traversability analysis on it. Meng et al. \cite{meng2023terrainnet} train a network to predict BEV-space semantics and terrain information from stereo images. Similarly to Gan et al. \cite{gan2022multitask}, ground truth is generated by aggregating pointclouds over time and hand-labeling the resulting output. They then demonstrate this network's ability to produce cost maps for planning by hand-designing a cost function that uses the predicted semantic and terrain information. 

We differentiate ourselves from prior work by leveraging high-precision laser scanners to generate high-quality ground truth environment models. This gives us a more accurate and complete pointcloud of the environment to leverage for learning. In fact, the quality of our environment representation is sufficient for us to simulate realistic lidar scans for training. Additionally, we differentiate ourselves from prior work in that we predict a number of geometric features important to traversability with uncertainty estimates, as well as a principled way of leveraging this uncertainty in traversability estimation. This allows our method to be aware of uncertainty and robot-agnostic, while remaining label-free.  


\section{Methodology}

\begin{figure*}
    \centering
    \vspace{0.25cm}
    \includegraphics[width=0.75\linewidth]{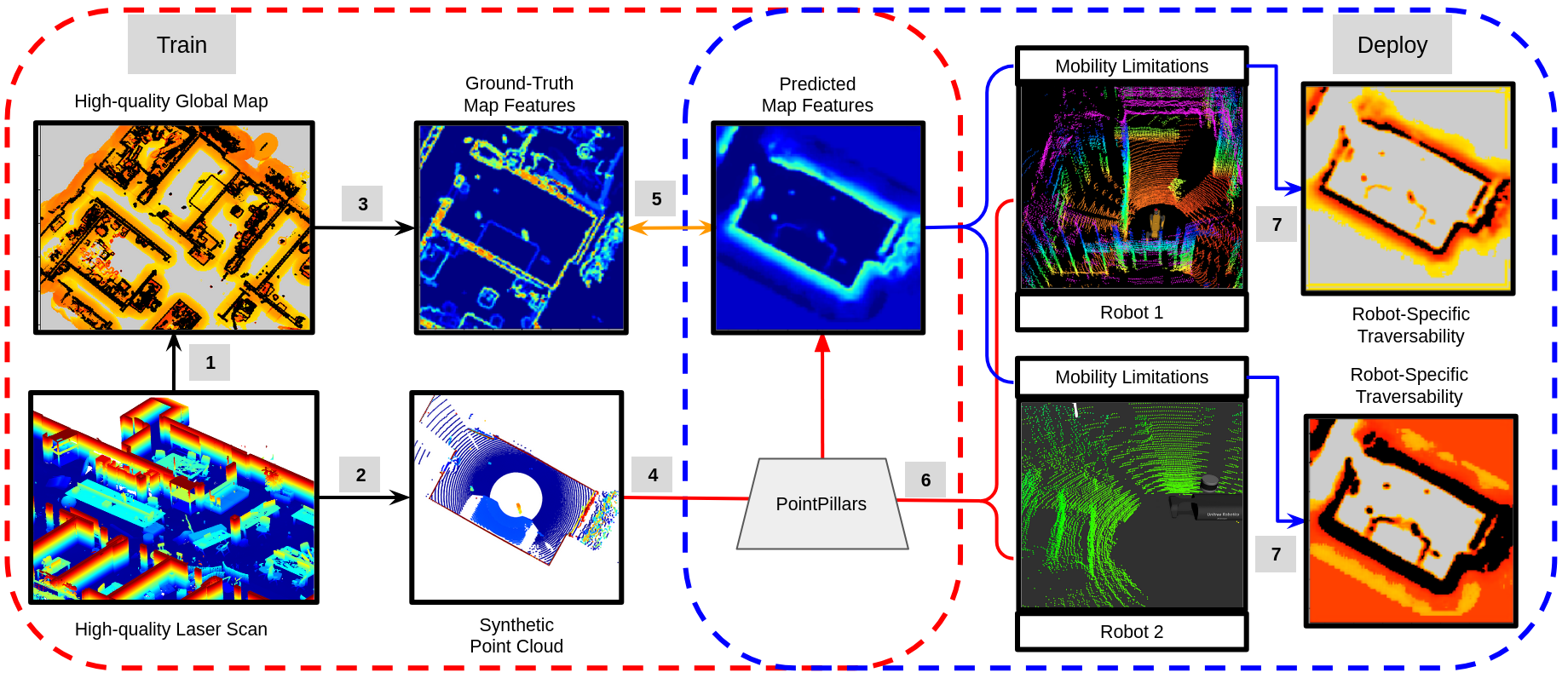}
    \caption{An overview of our algorithm. We first create a high-quality map (1) from a pointcloud from a laser scanner. We then generate synthetic pointclouds (2) and their corresponding ground-truth map features (3). We train UNRealNet to predict the high-quality, robot-agnostic map features from the noisy, synthetic pointclouds (4, 5). We can then deploy this network on multiple robots (6) by leveraging a robot-specific traversability function (7). Red arrows denote network inputs, blue networks denote network outputs, and the orange arrow denotes the training objective (Equation \ref{eq:loss_fn}).}
    \label{fig:algo_overview}
\end{figure*}

Our goal is to train a neural network to predict dense map of traversability features ($\mathcal{M} \in \mathbb{R}^{C \times W \times H}$, where $C$ is the number of features, $W$ is the map width, and $H$ is the map height) from a sparse, instantaneous pointcloud ($\mathcal{P} \in \mathbb{R}^{P \times 3}$, where $P$ is the number of points). Doing so will result in a local map that is quick to compute and is robust to occlusions as well as odometry drift. A high-level summary of our method is presented in Figure \ref{fig:algo_overview}.

\subsection{Creating a High-Quality, Real-World Dataset}

\subsubsection{Mapping}
To begin, we leverage a high-quality laser scanner (a FARO Focus and Leica BLK360) and their pointcloud registration software to generate a high-quality pointcloud of six locations, over four unique environments. This approach is preferred to relying on offline SLAM algorithms for the following reasons:

\begin{enumerate}
    \item With high precision and accuracy (up to 1 million points per scan, at 1.0mm precision) \cite{faro_datasheet}, the precision of these laser scanners exceeds that of common sensors in robotics such as lidar and camera.
    \item The high density of points in these scans makes it possible to generate high quality data from synthetic viewpoints. This allows us to create training pairs from places the robot did not (or could not) traverse.
    \item Near full coverage of the environment can be achieved by not being reliant on the traversability capabilities of the robot using the sensor.
\end{enumerate}

\begin{figure}
    \centering
    \includegraphics[width=\linewidth]{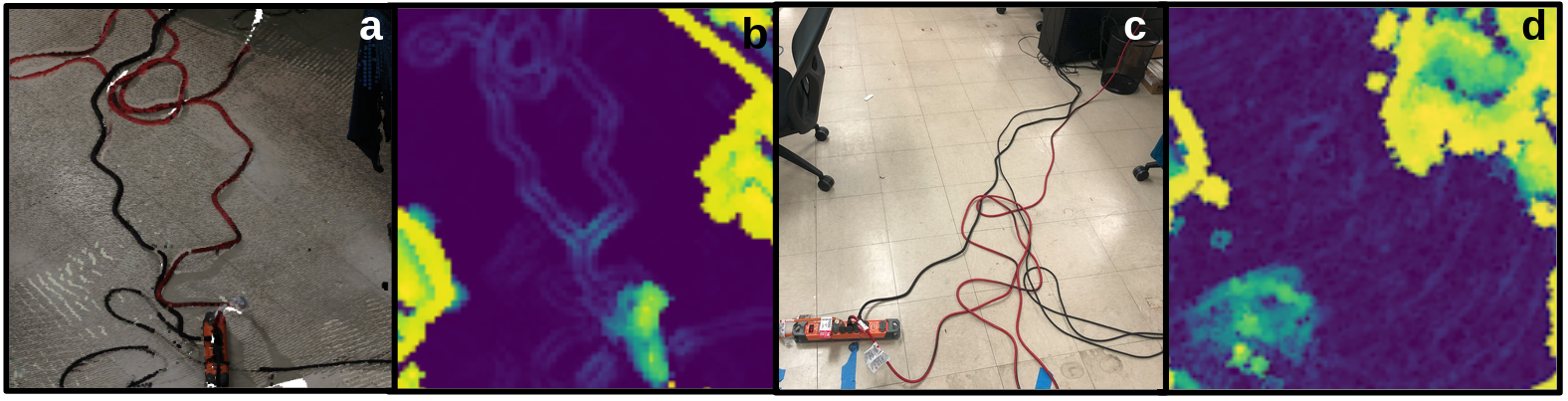}
    \caption{An example that highlights the value of the high-resolution scanner. (a) the pointcloud from the scanner. (b) the slope feature of the corresponding local map. We are able to identify the edges of each wire. (c) A photo of the environment and (d) the slope feature from the registered pointcloud. The noise from the lidar and SLAM is too high to see the wires.}
    \label{fig:wire_example}
    \vspace{-0.5cm}
\end{figure}

In order to generate high quality, metric-space traversability labels from this pointcloud, we run a simple elevation mapping and freespace detection algorithm. The resulting freespace was then manually refined. This step can easily be replaced with a more sophisticated algorithm \cite{zhang2016easy, velas2018cnn} if needed. Given this elevation map, we can then construct high-quality traversability features. As is common in prior work \cite{chilian2009stereo, fankhauser2018robust}, we use the step, slope and roughness of a local neighborhood of elevation cells (the same as Chilian et al. \cite{chilian2009stereo}). We find that the high precision of the pointcloud allows for relatively nuanced costing of subtle terrain features such as loose wires and smaller changes in elevation (see Figure \ref{fig:wire_example}). The end result of this local mapping step is that we now have a dense, accurate pointcloud $\mathcal{P}_{gt}$ and corresponding map features $\mathcal{M}_{gt}$ which can be leveraged for training. 

\subsubsection{Synthetic pointclouds}
A key advantage of using a dense pointcloud of the environment is that it is possible to generate high-quality synthetic data from many viewpoints that were not directly experienced by the robot. Thus, in order to generate training pairs $(\mathcal{P}, \mathcal{M})$ from $\mathcal{M}_{gt}, \mathcal{P}_{gt}$ to train our network, we can sample a pose in freespace $p$ and project $\mathcal{P}_{gt}$ into the frame defined by $p$ (equivalent to the CP method described by Langer et al. \cite{langer2020domain}). This gives us a pointcloud $\mathcal{P}$ that resembles what the sensor would have observed from pose $p$. In order to generate the corresponding set of map features, we can transform and crop the global map based on $p$. Finally, we can apply a noising pipeline (including range noise, robot self-hit masks, etc.) to make $\mathcal{P}$ resemble the output of a robot sensor \cite{ouster_datasheet}. 

\begin{figure*}
    \centering
    \vspace{0.25cm}
    \includegraphics[width=0.95\linewidth]{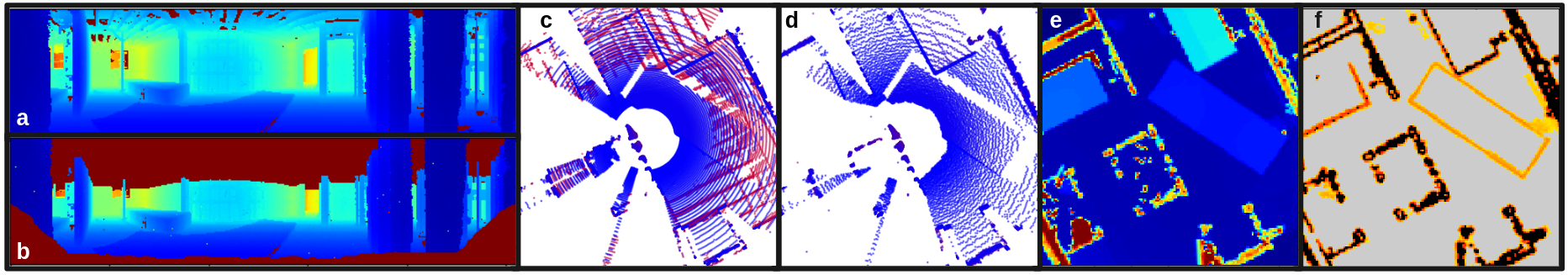}
    \caption{A visualization of a datapoint in our dataset, showing (a) the depth image obtained from simulating lidar in the ground-truth pointcloud (blue=close, red=far), (b) the depth image obtained by running the original depth image through the noising pipeline, (c) the BEV projection of the original pointcloud, (d) the BEV projection of noised pointcloud, and corresponding crops from the ground-truth elevation map (e) and a sample traversability map (f).}
    \label{fig:simulated}
\end{figure*}

\subsection{Traversability Prediction with UNRealNet}
UNRealNet is a PointPillars-based \cite{lang2019pointpillars} network designed to predict traversability features directly from the above synthetic pointclouds. Similarly to PointPillars, we first define a metric region around the ego-pose (consisting of a resolution, width and height). All points in this region are passed through a PointPillars network (consisting of a small PointNet \cite{qi2017pointnet}, cell-wise max-pooling, and then a UNet \cite{ronneberger2015u}, which produces a cell-wise, factorized Gaussian distribution for several traversability features (Equation \ref{eq:net_output}). This network is supervised by minimizing the negative log-likelihood of the ground-truth map crop under the predicted map distribution, where unobserved cells in the ground truth map crop are masked out via a mask $M_{gt}$ (Equation \ref{eq:loss_fn}). Note that the network receives a training signal for cells observed in the label that are not observed in the input, thus requiring the network to inpaint map features.

\vspace{-0.4cm}
\begin{equation}
    \label{eq:net_output}
    f_\theta(\mathcal{P}) = \mathcal{N}(\tilde{\mu}, \tilde{\sigma}), \quad \tilde{\mu} \in \mathbb{R}^{C \times W \times H}, \quad \tilde{\sigma} \in \mathbb{R}^{C \times W \times H}
\end{equation}
\vspace{-0.5cm}

\begin{equation}
    \label{eq:loss_fn}
    \begin{split}
        \mathcal{L}(\hat{\mathcal{M}}, f_\theta(\mathcal{P})) & = \frac{M_{gt}}{\sum M_{gt}} \sum^{W}_{i=0} \sum^{H}_{j=0} \sum^{C}_{f=0} \bigg[ \\
        & -log(p(\hat{\mathcal{M}}^{f, i, j}, \mathcal{N}(\tilde{\mu}^{f, i, j}, \tilde{\sigma}^{f, i,j}))) \bigg]
    \end{split}
    \vspace{-0.4cm}
\end{equation}

\subsection{Probabilistic Traversability Function}
While it is possible to compute a single traversability function at train-time, this approach has a significant downside in that traversability estimates cannot be changed at deployment time. This limitation would require re-training of the network for any additional robot platform, or even when tuning the cost function to specific user needs or risk tolerances. As such, it is desirable to be able to re-compute traversability using the predicted features from the network.

It is common practice to use Equation \ref{eq:fank_trav} as a cost function for planning, where the features in question are geometric terrain properties such as local slope and roughness. However, given our uncertainty estimates, it is more intuitive to treat traversability as the probability that all map features are below their critical values (Equation \ref{eq:prob_trav}).

\begin{equation}
    \label{eq:prob_trav}
    p(T^{i, j}) = p(T^{i, j} < f_{crit}, \forall f)
\end{equation}

Assuming that each map feature is an independent Gaussian (i.e. $f_{i, j} = (\mu_{i, j}, \sigma_{i, j})$) allows us to simplify Equation \ref{eq:prob_trav} into Equation \ref{eq:prob_trav2}. This formulation only requires the evaluation and multiplication of several Gaussian cdfs, which are easily computable at the speeds required for multi-Hz traversability estimation over maps with tens of thousands of cells. Additionally, this simplifies the burden on the robot operator through the elimination of the $\alpha$ parameter.

\begin{equation}
    \label{eq:prob_trav2}
    p(T^{i,j}) = \prod_f cdf(\mathcal{N}(\mu^{f, i, j}, \sigma^{f, i, j}), f_{crit})
\end{equation}

\section{Experiments, Results and Analysis}
In total, we collected high-quality laser scans of five environments at three active construction sites. Each scan consists of a single floor (two floors were collected at two sites). The above data generation process was run on each environment, yielding a total of around 30,000 train samples and 12,000 test samples (roughly a $70-30$ test-train split). Additionally, a sixth environment (at a different site) was collected and used solely for testing. This site yielded an additional 7,000 test samples. UNRealNet was trained to predict $7m \times 7m$ local maps at a resolution of $5cm$, though maps of arbitrary size can be predicted using this pipeline. Our traversability features consisted of the following:

\begin{enumerate}
    \item \textbf{terrain}: 1st percentile of heights in a cell.
    \item \textbf{elevation}: 99th percentile of heights in a cell, after removing points over a certain height.
    \item \textbf{step}: difference between elevation and terrain.
    \item \textbf{local slope}: slope in a foothold-sized radius.
    \item \textbf{local rough}: height variance in a foothold-sized radius.
    \item \textbf{slope}: slope in a robot footprint-sized radius.
    \item \textbf{rough}: height variance in a robot footprint radius.
\end{enumerate}
Slope, step and roughness were computed in the same way as Fankhauser et al. \cite{fankhauser2018robust}. In total, this resulted in a $7 \times 140 \times 140$ tensor of labels per datapoint.

\subsection{Network Performance}

In order to evaluate UNRealNet we ask the following:

\begin{enumerate}
    \item Does UNRealNet outperform standard baselines?
    \item Does the probabilistic traversability function better capture ground-truth traversability?
\end{enumerate}

\subsubsection{Does UNRealNet Outperform Standard Baselines?}

We first examine UNRealNet's ability to produce traversability features from simulated lidar scans. In order to quantify this, we use the ground-truth map features for datapoints in our test set. Our primary baseline is the mapping algorithm used to generate features from the aggregated laser scans, but given the simulated pointcloud used as input. This baseline is chosen because it a perfectly observed pointcloud run through this baseline would generate zero error. Thus it serves as a good measure of how well each method can mitigate the effect of partial sensing.

Since the baseline local mapping method does not produce uncertainty estimates, we report RMSE between the mean of the predicted distribution and the ground-truth features for networks. We compare the following ablations:
\begin{enumerate}
    \item (baseline): The local mapping method used to generate ground-truth features, applied to the synthetic pointcloud. Empty cells are filled with a value of zero.
    
    \item (Telea): baseline, but leverages the inpainting method proposed by Telea et al. \cite{telea2004image} for unknown cells.
    
    \item (NS): baseline, but leverages the inpainting method based on Navier-Stokes proposed by Bertalmio et al. \cite{bertalmio2001navier} for unknown cells.
    
    \item (oracle): Equivalent to baseline, except the placeholder is determined by computing the RMSE-minimizing value for each channel (using the ground-truth).
    
    \item (no aug): An ablation with the omission of the pointcloud augmentation pipeline. This ablation is provided to quantify the importance of making the simulated lidar reflect real sensing.
\end{enumerate}
We then evaluate each method via the following metrics:
\begin{enumerate}
    \item (both) avg. RMSE for cells with ground-truth values. 
    \item (observed) avg. RMSE for cells observed in the input pointcloud. This captures each method's ability to correct for occlusion-induced mapping errors.
    \item (inpaint) avg. RMSE for cells which are not observed in the pointcloud, but have ground-truth values. This captures the ability of each method to interpolate and extrapolate known information to unknown areas.
\end{enumerate}

Results for synthetic scans on a spatially held-out region of each environment are reported in Table \ref{tab:rmse_results}. We also report results for the sixth environment (env6, in red to denote that it was not in the train set) to demonstrate UNRealNet's ability to generalize to novel environments. We also report results from a robot traversal of env6 with the on-board lidar (env6, ouster). To get proper correspondences with the ground-truth map, we first align ground-truth cloud and the aggregated pointcloud via ground-plane matching and ICP. Since there is some drift in the SLAM solution, we further refine each individual scan by performing another ICP step for each individual scan. In total, this yields around 800 scans.

Standard deviations are reported over three seeds for \textit{net} and \textit{no aug}. Note that the augmentation pipeline introduces a small amount of variance in the baselines, which are otherwise deterministic. Also note that the RMSE on observed cells for all non-learned methods is nearly identical as they copy the values of observed cells.

\begin{table}[]
    \centering
    \tiny
    \vspace{0.25cm}
    \begin{tabular}{c c||c|c|c}
        Env & Method & RMSE (observed) & RMSE (inpaint) & RMSE (both) \\
        \hline
        
        env1& net (ours)    & $\mathbf{0.1562 \pm 0.0001}$  & $\mathbf{0.1987 \pm 0.0009}$  & $\mathbf{0.1849 \pm 0.0006}$ \\
                 & no aug    & $0.2117 \pm 0.0101$  & $0.2618 \pm 0.0121$  & $0.2453 \pm 0.0113$ \\
                 & oracle    & $0.2817 \pm 0.0003$  & $0.2736 \pm 0.0003$  & $0.2779 \pm 0.0003$ \\ 
                 & telea     & $0.2817 \pm 0.0003$  & $2.0402 \pm 0.0017$  & $1.6424 \pm 0.0029$ \\ 
                 & ns        & $0.2817 \pm 0.0003$  & $0.3502 \pm 0.0003$  & $0.3298 \pm 0.0003$ \\ 
                 & baseline  & $0.2817 \pm 0.0003$  & $0.4097 \pm 0.0007$  & $0.3740 \pm 0.0005$ \\ 
               \hline
               
         env2  & net (ours)    & $\mathbf{0.2718 \pm 0.0030}$  & $\mathbf{0.3608 \pm 0.0036}$  & $\mathbf{0.3284 \pm 0.0035}$ \\ 
                     & no aug    & $0.3313 \pm 0.0141$  & $0.4208 \pm 0.0142$  & $0.3877 \pm 0.0137$ \\ 
                     & oracle    & $0.3731 \pm 0.0016$  & $0.4210 \pm 0.0007$  & $0.4080 \pm 0.0011$ \\ 
                     & telea     & $0.3731 \pm 0.0016$  & $2.0930 \pm 0.0154$  & $1.6450 \pm 0.0153$ \\ 
                     & ns        & $0.3731 \pm 0.0016$  & $0.4638 \pm 0.0015$  & $0.4327 \pm 0.0017$ \\ 
                     & baseline  & $0.3731 \pm 0.0016$  & $0.4659 \pm 0.0007$  & $0.4350 \pm 0.0012$ \\
               \hline
               
         env3 & net (ours)    & $\mathbf{0.1705 \pm 0.0021}$  & $\mathbf{0.2370 \pm 0.0017}$  & $\mathbf{0.2172 \pm 0.0017}$ \\ 
                     & no aug    & $0.2319 \pm 0.0127$  & $0.2902 \pm 0.0114$  & $0.2724 \pm 0.0111$ \\
                     & oracle    & $0.2669 \pm 0.0005$  & $0.2937 \pm 0.0004$  & $0.2883 \pm 0.0004$ \\ 
                     & telea     & $0.2669 \pm 0.0005$  & $2.3610 \pm 0.0058$  & $1.9318 \pm 0.0064$ \\ 
                     & ns        & $0.2669 \pm 0.0005$  & $0.3769 \pm 0.0003$  & $0.3474 \pm 0.0003$ \\ 
                     & baseline  & $0.2669 \pm 0.0005$  & $0.4370 \pm 0.0002$  & $0.3940 \pm 0.0002$ \\
               \hline
               
         env4& net (ours)    & $\mathbf{0.1413 \pm 0.0008}$  & $\mathbf{0.2252 \pm 0.0014}$  & $\mathbf{0.2071 \pm 0.0012}$ \\
                     & no aug    & $0.2033 \pm 0.0099$  & $0.2845 \pm 0.0077$  & $0.2663 \pm 0.0078$ \\
                     & oracle    & $0.2629 \pm 0.0001$  & $0.2739 \pm 0.0003$  & $0.2728 \pm 0.0002$ \\ 
                     & telea     & $0.2629 \pm 0.0001$  & $2.8353 \pm 0.0028$  & $2.4492 \pm 0.0017$ \\ 
                     & ns        & $0.2629 \pm 0.0001$  & $0.4132 \pm 0.0005$  & $0.3830 \pm 0.0005$ \\ 
                     & baseline  & $0.2629 \pm 0.0001$  & $0.4165 \pm 0.0009$  & $0.3868 \pm 0.0007$ \\ 
               \hline
               
         env5 & net (ours)    & $\mathbf{0.1748 \pm 0.0016}$  & $\mathbf{0.2792 \pm 0.0019}$  & $\mathbf{0.2483 \pm 0.0016}$ \\ 
                     & no aug    & $0.2313 \pm 0.0104$  & $0.3260 \pm 0.0135$  & $0.2969 \pm 0.0114$ \\ 
                     & oracle    & $0.2439 \pm 0.0005$  & $0.3345 \pm 0.0005$  & $0.3084 \pm 0.0002$ \\ 
                     & telea     & $0.2439 \pm 0.0005$  & $2.4452 \pm 0.0238$  & $1.9733 \pm 0.0227$ \\ 
                     & ns        & $0.2439 \pm 0.0005$  & $0.3635 \pm 0.0008$  & $0.3297 \pm 0.0008$ \\ 
                     & baseline  & $0.2439 \pm 0.0005$  & $0.4263 \pm 0.0005$  & $0.3761 \pm 0.0008$ \\
                \hline

        \color{red} env6   & net (ours)    & $\mathbf{0.1920 \pm 0.0012}$  & $\mathbf{0.2708 \pm 0.0016}$  & $\mathbf{0.2537 \pm 0.0015}$ \\ 
                             & no aug    & $0.2466 \pm 0.0087$  & $0.3156 \pm 0.0086$  & $0.3001 \pm 0.0079$ \\
                             & oracle    & $0.2985 \pm 0.0004$  & $0.2998 \pm 0.0003$  & $0.3023 \pm 0.0003$ \\ 
                             & telea     & $0.2985 \pm 0.0004$  & $2.8835 \pm 0.0098$  & $2.4849 \pm 0.0099$ \\ 
                             & ns        & $0.2985 \pm 0.0004$  & $0.4364 \pm 0.0005$  & $0.4092 \pm 0.0005$ \\ 
                             & baseline  & $0.2985 \pm 0.0004$  & $0.4598 \pm 0.0008$  & $0.4260 \pm 0.0008$ \\
        \hline

        \color{red} env6,    & net (ours)    & $\mathbf{0.2200 \pm 0.0014}$  & $\mathbf{0.2666 \pm 0.0006}$  & $\mathbf{0.2567 \pm 0.0007}$ \\ 
        \color{red} ouster   & no aug    & $0.2217 \pm 0.0018$  & $0.2744 \pm 0.0014$  & $0.2628 \pm 0.0014$ \\
                             & oracle    & $0.3229 \pm 0.0007$  & $0.2941 \pm 0.0000$  & $0.3046 \pm 0.0002$ \\ 
                             & telea     & $0.3229 \pm 0.0007$  & $2.4206 \pm 0.0213$  & $2.1006 \pm 0.0209$ \\ 
                             & ns        & $0.3229 \pm 0.0007$  & $0.4322 \pm 0.0020$  & $0.4118 \pm 0.0017$ \\ 
                             & baseline  & $0.3229 \pm 0.0007$  & $0.4658 \pm 0.0003$  & $0.4374 \pm 0.0004$ \\ 
    \end{tabular}
    \caption{Results for local mapping accuracy (env6 colored red to signify that it is a held-out location)}
    \label{tab:rmse_results}
\end{table}

Overall, we observe that UNRealNet outperforms all baselines, for all metrics, on all evaluation datasets. Interestingly, we observe that this performance gain is fairly similar for both observed and unobserved cells. This is likely due to factors such as occlusion and sparse sensing resulting in mis-estimation of terrain features (e.g. only observing the top part of an object and mistaking it for the ground). Furthermore, we observe that both inpainting baselines perform quite poorly, especially when compared to the oracular baseline. This is likely due to the relative sparsity of the input data (we only have returns for a relatively low fraction of cells, compared to traditional CV datasets for inpainting). Furthermore, data is sparser towards the edges of the map, meaning that extrapolation, in addition to interpolation, is necessary to achieve low RMSE. Finally, we observe that adding augmentations similar to the actual robot payload improves overall performance.

\subsubsection{Does the Probabilistic Traversability Function Better Capture Ground-Truth?}

In order to evaluate the efficacy of the probabilistic traversability function, we compare the accuracy to a ground-truth traversability derived from crops of the global map. For each data point, a random set of traversability threshods was generated. These traversability thresholds were used to generate a traversability map by checking if any ground-truth features exceeded their corresponding threshold, per cell. We then evaluated the proposed traversability function (Prob.) and the one used by Chilian and Fankhauser \cite{chilian2009stereo, fankhauser2018robust} (Det.), on the predicted traversability features from the network. We report the MAE of the predicted traversability value to the ground truth map. Results are presented in Table \ref{tab:prob_trav} for the test set, over a random sampling of traversability thresholds. 


\begin{table}[]
    \centering
    \vspace{0.25cm}
    \begin{tabular}{c|c}
         Method & MAE \\
         \hline
         Det. \cite{chilian2009stereo, fankhauser2016universal} & $0.3000 \pm 0.0546$ \\
         Prob. (ours) & $\mathbf{0.2326 \pm 0.0694}$ \\
    \end{tabular}
    \caption{Probabilistic traversability experiment results.}
    \label{tab:prob_trav}
\end{table}

\begin{figure}
    \centering
    \includegraphics[width=0.85\linewidth]{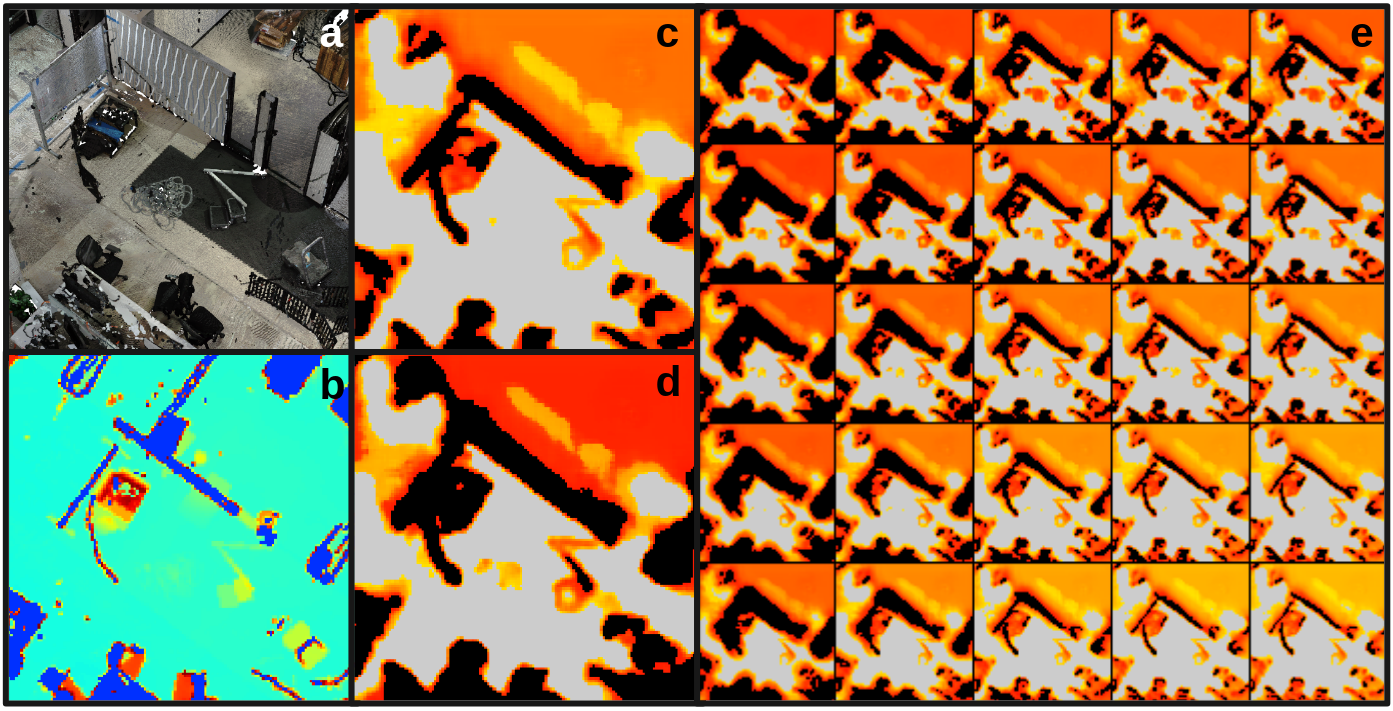}
    \caption{Traversability predictions using the same network output on a sample in env6. (a) A crop from the FARO-scanned environment. There is a pile of wires in the middle. (b) The ground-truth elevation map. (c) The traversability map generated from parameters for a Spot. (d) The traversability map generated from parameters for an AlienGo. (e) A mosaic of traversability maps by varying the local slope (y axis) and robot slope (x axis) thresholds.}
    \label{fig:trav_interp}
    \vspace{-0.5cm}
\end{figure}

Overall, we observe that the uncertainty-aware traversability function is closer to the ground-truth traversability. This is likely because incorporating uncertainty allows the traversability estimation to be less sensitive to mispredictions of features (due to assigning high uncertainty to said areas).


\subsection{Hardware Experiments}

We also ran a number of experiments on multiple legged platforms to demonstrate the efficacy of our method in practice. Our traversability pipeline was run on both a Boston Dynamics Spot and Unitree Aliengo, each with a custom payload similar to Bouman et al. \cite{bouman2020autonomous}. 

\begin{figure*}
    \centering
    \vspace{0.25cm}
    \includegraphics[width=0.9\linewidth]{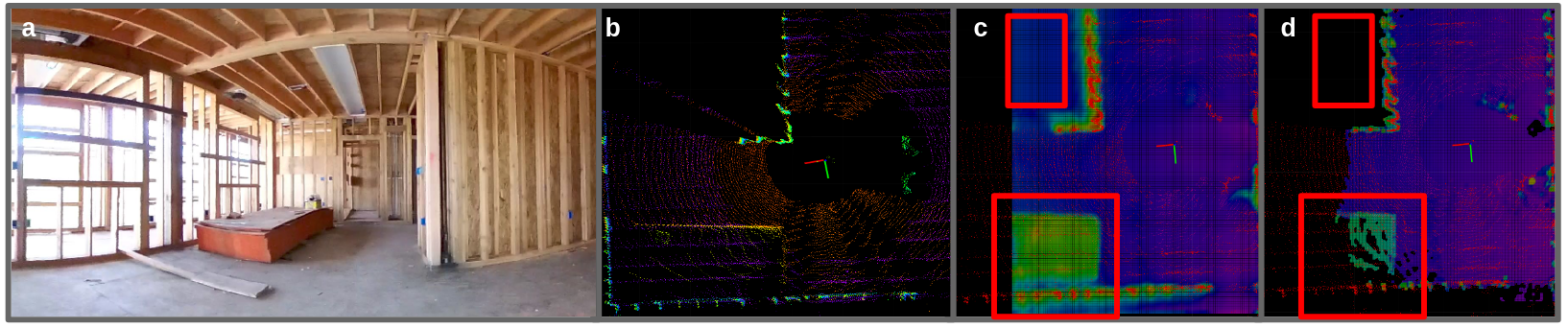}
    \caption{Case study for elevation prediction. (a) An FPV image of the scene. (b) The input pointcloud (red=low, purple=high) (c) The learned, Kalman-filtered elevation map from UNRealNet (purple=low, red=high) has inpainted the floor behind the wall, as well as the construction material and pillars. (d) Elevation map from STEP \cite{fan2021step} is incomplete due to occlusions.}
    \label{fig:torrance_case_study}
\end{figure*}

\begin{figure}
    \centering
    \includegraphics[width=0.8\linewidth]{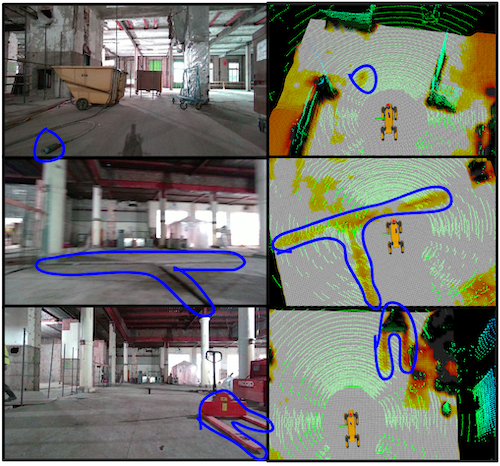}
    \caption{Additional qualitative results from a seventh environment. We are able to capture detailed traversability estimates including small plugs (top), shallow negative obstacles (middle) and complex geometry (bottom). Correspondences between FPV and BEV are circled in blue. FPV viewpoints were taken at different times than BEV to keep the object of interest in the field of view.}
    \label{fig:salt_lake_qual}
    \vspace{-0.75cm}
\end{figure}


\subsubsection{Robot-Agnosticism}
Presented in Figure \ref{fig:trav_interp} are multiple traversability maps generated from the same UNRealNet prediction from env 6. Our traversability function is able to take into account robot-specific mobility limitations and produce a corresponding cost. Of note is that UNRealNet is able to detect and evaluate traversability of a small pile of wires, which becomes more costly the more stringent the traversability thresholds become.

\subsubsection{Kalman Filter}

When deploying on the robot, we can fuse multiple predictions together to get a higher-quality map (at the cost of becoming reliant on accurate odometry). This is accomplished by leveraging the uncertainty in the neural network predictions ($\mu, \sigma$, per feature, per cell) to run a one-dimensional Kalman Filter (Equation \ref{eq:kalman}, where $\hat{\mu}, \hat{\sigma}$ are the running estimates of the Kalman filter) \cite{fankhauser2018probabilistic}. Traversability $\hat{T}^{i, j}$ can then be computed from the Kalman-filtered estimates $(\hat{\mu}, \hat{\sigma})$. We observe that the Kalman filter generally improves the estimation of terrain features. 

\begin{equation}
    \label{eq:kalman}
    \hat{\mu}' = \frac{\hat{\mu} \sigma^2 + \mu \hat{\sigma}^2}{\hat{\sigma}^2 + \sigma^2}, \quad
    \hat{\sigma}' = \sqrt{\frac{\hat{\sigma}^2 \sigma^2}{\hat{\sigma}^2 + \sigma^2}}
\end{equation}



\subsubsection{Inference Speed}
We first report the speed of our method in comparison to STEP \cite{fan2021step}.  Both methods were run with an AMD Ryzen 9 5900hx CPU and NVIDIA 3060 Laptop GPU. We observe that although slower, the runtime of our method is comparable to STEP (around $7hz$ compared to STEP's $13hz$, and $10hz$ rate of pointclouds from the lidar). The majority of the runtime in our method is consumed by preprocessing the pointcloud for the network, and converting the network's output into a grid map. We argue that much of this runtime can be reduced by re-implementing our Python inference node in C++, and leveraging tools such as TensorRT \cite{tensor_rt}. Additionally, we note that $7hz$ inference is fast enough for legged platforms with a top speed of under $2m/s$, though this may not be sufficient for faster platforms.

\subsubsection{Case Studies}

One particular case study is presented in Figure \ref{fig:torrance_case_study}. We compare the learned elevation map to the elevation map produced by STEP. We can observe that UNRealNet has learned to extrapolate using environmental priors, in that it successfully inpaints the crate, pillars, and the floor in between. Additionally, it correctly predicts the floor behind the wall. We also provide additional qualitative results from a seventh environment (for which we didn't have a ground-truth pointcloud) in Figure \ref{fig:salt_lake_qual} that highlight additional nuances in our traversability estimation.

\section{Conclusion and Future Work}

We presented a method for learning traversability information by generating synthetic sensing from high-quality scans of real-world environments. We demonstrate that this method results in improved local mapping performance and demonstrate its efficacy for traversability estimation on hardware. 

Future work includes generating better training labels, such as correspondences between observed terrain features and proprioception \cite{castro2022does, cai2022risk, frey2023fast} or demonstrations \cite{bagnell2010learning, wigness2018robot, triest2023learning}. The work done by Meng et al. \cite{meng2023terrainnet} provides a good roadmap for how to incorporate semantics into a similar mapping pipeline. Additionally, datasets such as Matterport 3d \cite{chang2017matterport3d} and TartanAir \cite{wang2020tartanair} can be leveraged to obtain joint geometric and semantic labels without any additional human labeling (although fine-tuning on real data may be required). Lastly, combining UNRealNet with object detection and tracking algorithms would allow for increased performance in the presence of dynamic obstacles.

\section{Acknowledgements}

The authors thank the rest of the Field AI team for their help in collecting data and testing on hardware.

{
    \bibliographystyle{IEEEtran}
    \bibliography{refs}
}

\newpage


\end{document}


\appendix

\subsection{Network Architecture}

The specific architecture used for our network is described in Table \ref{tab:net_arch}.

\begin{table*}[]
    \centering
    \begin{tabular}{c||c|c|c|c}
        Layer & Input Dim & Output Dim & Kernel & Activation  \\
        \hline
        PointNet & $140 \times 140 \times P \times 8$ & $32 \times 140 \times 140$ & - & ReLU \\
        Downsample 1 & $32 \times 140 \times 140$ & $32 \times 70 \times 70$ & $3 \times 3$ & ReLU \\
        Downsample 2 & $32 \times 70 \times 70$ & $64 \times 35 \times 35$ & $3 \times 3$ & ReLU \\
        Downsample 3 & $64 \times 35 \times 35$ & $128 \times 17 \times 17$ & $3 \times 3$ & ReLU \\
        Downsample 4 & $128 \times 17 \times 17$ & $256 \times 8 \times 8$ & $3 \times 3$ & ReLU \\
        Channel-Wise Conv & $256 \times 8 \times 8$ & $256 \times 8 \times 8$ & - & - \\
        Upsample 1 & $256 \times 8 \times 8$ & $128 \times 17 \times 17$ & $3 \times 3$ & ReLU \\
        Upsample 2 & $128 \times 17 \times 17$ & $64 \times 35 \times 35$ & $3 \times 3$ & ReLU \\
        Upsample 3 & $64 \times 35 \times 35$ & $32 \times 70 \times 70$ & $3 \times 3$ & ReLU \\
        Upsample 4 & $32 \times 70 \times 70$ & $32 \times 140 \times 140$ & $3 \times 3$ & ReLU \\
        Mean Feature Head & $32 \times 140 \times 140$ & $8 \times 140 \times 140$ & $3 \times 3$ & - \\
        Stddev Feature Head & $32 \times 140 \times 140$ & $8 \times 140 \times 140$ & $3 \times 3$ & Exp \\
    \end{tabular}
    \caption{Network Architecture}
    \label{tab:net_arch}
\end{table*}

\subsection{Noising Pipeline}
Details for the noising pipeline are provided in Table \ref{tab:noising_params}. The pipeline is instantiated as follows: 
\begin{enumerate}
    \item \textbf{Ceiling Crop: } remove all depth measurements that are $\Delta z_{max}$ above the ego-height.
    \item \textbf{Robot Mask: } remove all depth measurements that coincide with an existing mask corresponding to the body of the robot.
    \item \textbf{Salt and Pepper Noise: } Apply Equation \ref{eq:salt_pepper_noise} to each range measurement $r$.
    \item \textbf{Range Noise: } Apply Equation \ref{eq:linear_noise_model} to each range measurement $r$.
\end{enumerate}

\begin{equation}
    \label{eq:salt_pepper_noise}
    r = \begin{cases}
        r & \hbox{with probablility } p_r\\
        \mathcal{U}(0, r_{max}) & \hbox{otherwise}
    \end{cases}
\end{equation}

\begin{equation}
    \label{eq:linear_noise_model}
    r = r + \sigma(r), \quad \sigma(r) \sim \mathcal{N}(0, mr + c)
\end{equation}

Lidar masks are randomly sampled from a set of four existing masks for existing legged robots and payloads. Each step in the pipeline was applied with some associated probability and has hyperparameters that are randomly sampled within a given range.

\begin{table}[]
    \centering
    \begin{tabular}{c|c|c}
        Processing Step & Parameter & Value/Range \\
        \hline
        Ceiling Crop & $p$         & 0.5 \\
                     & $\Delta z_{max}$ & [0.5, 1.0] \\
        \hline
        Robot Mask   & $p$         & 0.8   \\
                     & N masks     & 4     \\
        \hline
        Salt and     & $p$         & 1.0   \\
        Pepper Noise & $p_r$       & 0.999 \\
                     & $r_{max}$   & 10.0  \\
        \hline
        Range Noise  & $p$         & 1.0   \\
                     & $c$         & 0.0 \\
                     & $m $        & [0.001, 0.01]  \\
        
    \end{tabular}
    \caption{List of parameters for the noising pipeline. Stages were applied with probability $p$, and ranges were uniformly sampled.}
    \label{tab:noising_params}
\end{table}

\subsection{Timing Information}
We first report the speed of our method in comparison to STEP \cite{fan2021step}. Results are presented in Table \ref{tab:timing_info}. Both methods were run with an AMD Ryzen 9 5900hx CPU and NVIDIA 3060 Laptop GPU.

\begin{table}[]
    \centering
    \begin{tabular}{c||c c|c c}
        Method & Step & Time (ms) & Sub-step & Time (ms) \\
        \hline
        \hline
        STEP & Mapping & $41.95 \pm 10.97 $ & &  \\
             & Trav.   & $36.38 \pm 8.16 $ & &  \\
             & \textbf{Total}   & $\mathbf{78.34 \pm 13.67} $ & &  \\
        \hline
        Ours & Mapping & $135.16 \pm 49.76$ & Preprocess & $75.82 \pm 36.57$ \\
             &         &                    & Inference  & $29.09 \pm 23.78$ \\
             &         &                    & Serialize  & $30.25 \pm 19.22$ \\
             & Trav.   & $4.47   \pm 1.56$  &            &                   \\
             & \textbf{Total}   & $139.63 \pm 49.79$ &            &                   \\
    \end{tabular}
    \caption{Speed of our method compared to STEP.}
    \label{tab:timing_info}
    \vspace{-0.5cm}
\end{table}

\subsection{Qualitative Descriptions of Test Environments}
We give a general qualitative description of the six environments used in our experiments below. All environments contain a significant degree of clutter.

\begin{enumerate}
    \item \textbf{Env 1}: A floor from a large construction site. This environment contains many smaller rooms and walls, as well as non-rectilinear floorplans.

    \item \textbf{Env 2}: The ground floor from a construction site. This environment contains both indoor rooms and outdoor uneven terrain.

    \item \textbf{Env 3}: The third floor of the site from Env 2. This contains several large rooms connected by hallways, and various heights of floor.

    \item \textbf{Env 4}: The second floor of a third site. This environment contains enclosed rooms and large piles of construction material.

    \item \textbf{Env 5}: The ground floor of Env 4. This environment is largely open, with large piles of material.

    \item \textbf{Env 6}: An office environmnet. This environment contains the highest degree of clutter.
\end{enumerate}

